\documentclass{article} %
\usepackage[preprint]{colm2025_conference}

\usepackage{microtype}
\usepackage{hyperref}
\usepackage{url}
\usepackage{booktabs}

\usepackage{lineno}

\usepackage{xspace}
\usepackage{graphicx}
\usepackage{marvosym}

\definecolor{darkblue}{rgb}{0, 0, 0.5}
\definecolor{darkgreen}{RGB}{50,100,0}
\definecolor{darkred}{RGB}{200, 0, 0}
\definecolor{lightblue}{RGB}{220,235,250}
\hypersetup{colorlinks=true, citecolor=darkblue, linkcolor=darkblue, urlcolor=darkblue}

\usepackage{amsmath}
\usepackage{subcaption}
\usepackage{graphicx}
\usepackage{changes}
\usepackage{amsmath,mathtools}
\usepackage{changes}
\usepackage{pifont}

\usepackage{lipsum}
\usepackage{colortbl}
\usepackage{wrapfig}
\usepackage{xspace}
\usepackage{multirow}
\usepackage{enumitem}

\usepackage{pifont}
\usepackage{listings}
\usepackage{fontawesome}

\usepackage{wrapfig}
\usepackage{caption}
\usepackage{mdframed}

\usepackage{makecell}

\usepackage{float}

\newcommand{\lstbg}[3][0pt]{{\fboxsep#1\colorbox{#2}{\strut #3}}}

\lstdefinelanguage{diff}{
  basicstyle=\ttfamily\small,
  morecomment=[f][\lstbg{red!20}]-,
  morecomment=[f][\lstbg{green!20}]+,
}
\lstdefinelanguage{diffpython}{
  language=diff,
  morekeywords={def, if, else, for, while, return, import, from, as, class, with, try, except, finally, raise, lambda, and, or, not, in, is, None, True, False},
  morecomment=[l]{\#},
  morestring=[b]",
  morestring=[b]',
}
\newcommand{\xmark}{\textcolor{darkred}{\ding{55}}}

\usepackage{cleveref}
\crefname{section}{Sec.}{Sec.}

\usepackage[most,skins,theorems]{tcolorbox}
\tcbset{
  takeawaysbox/.style={
    title=Takeaways,
    colback=lightblue!80,
    colframe=black,
    fonttitle=\bfseries\small,
    coltitle=white,
    colbacktitle=black,
    enhanced,
    attach boxed title to top left={xshift=2.5mm,yshift=-2.5mm},
    boxed title style={rounded corners, size=small, colframe=black, colback=black},
    width=\linewidth,
    arc=3.5mm
  }
}

\NewDocumentCommand{\ganqu}
{ mO{} }{\textcolor{blue}{\textsuperscript{\textit{ganqu}}\textsf{\textbf{\small[#1]}}}}

\NewDocumentCommand{\ybsun}
{ mO{} }{\textcolor{magenta}{\textsuperscript{\textit{youbang}}\textsf{\textbf{\small[#1]}}}}

\newcommand{\method}{\textbf{TTRL}\xspace}

\title{TTRL: Test-Time Reinforcement Learning}

\author{Yuxin Zuo\thanks{Equal Contribution. Kaiyan Zhang (zhang-ky22@mails.tsinghua.edu.cn) and Ganqu Cui lead the project.
$\dag$: Corresponding authors.}\hspace{4pt}$^{1,2}$\quad
Kaiyan Zhang$^{*1}$ \quad
Li Sheng$^{1,2}$ \quad
Shang Qu$^{1,2}$ \quad
Ganqu Cui$^2$ \\
\textbf{Xuekai Zhu$^1$ \quad
Haozhan Li$^{1,2}$ \quad
Yuchen Zhang$^{2}$ \quad
Xinwei Long$^{1}$} \\
\textbf{Ermo Hua$^{1}$ \quad
Biqing Qi$^2$ \quad
Youbang Sun$^1$ \quad
Zhiyuan Ma$^1$ \quad
Lifan Yuan$^1$} \\
\textbf{Ning Ding$^{\dag1,2}$ \quad
Bowen Zhou$^{\dag1,2}$} \\
$^1$Tsinghua University \quad $^2$Shanghai AI Lab\\
\\
\href{https://github.com/PRIME-RL/TTRL}{https://github.com/PRIME-RL/TTRL}
}

\begin{document}

\ifcolmsubmission
\linenumbers
\fi

\maketitle
\begin{abstract}

This paper investigates Reinforcement Learning (RL) on data \textit{without} explicit labels for reasoning tasks in Large Language Models (LLMs). 
The core challenge of the problem is reward estimation during inference while not having access to ground-truth information.
While this setting appears elusive, we find that common practices in Test-Time Scaling (TTS), such as majority voting, yield surprisingly effective rewards suitable for driving RL training.
In this work, we introduce Test-Time Reinforcement Learning (\method{}), a novel method for training LLMs using RL on unlabeled data. \method enables self-evolution of LLMs by utilizing the priors in the pre-trained models.
Our experiments demonstrate that \method{} consistently improves performance across a variety of tasks and models.
Notably, \method{} boosts the pass@1 performance of \texttt{Qwen-2.5-Math-7B} by approximately $\mathbf{211\%}$ on the AIME 2024 with only unlabeled test data.
Furthermore, although  \method{} is only supervised by the maj@n metric, \method has demonstrated performance to consistently surpass the upper limit of the initial model maj@n, and approach the performance of models trained directly on test data with ground-truth labels.
Our experimental findings validate the general effectiveness of \method across various tasks and highlight \method{}'s potential for broader tasks and domains.
\end{abstract}

\begin{figure}[h]
\centering
\includegraphics[width=\textwidth]{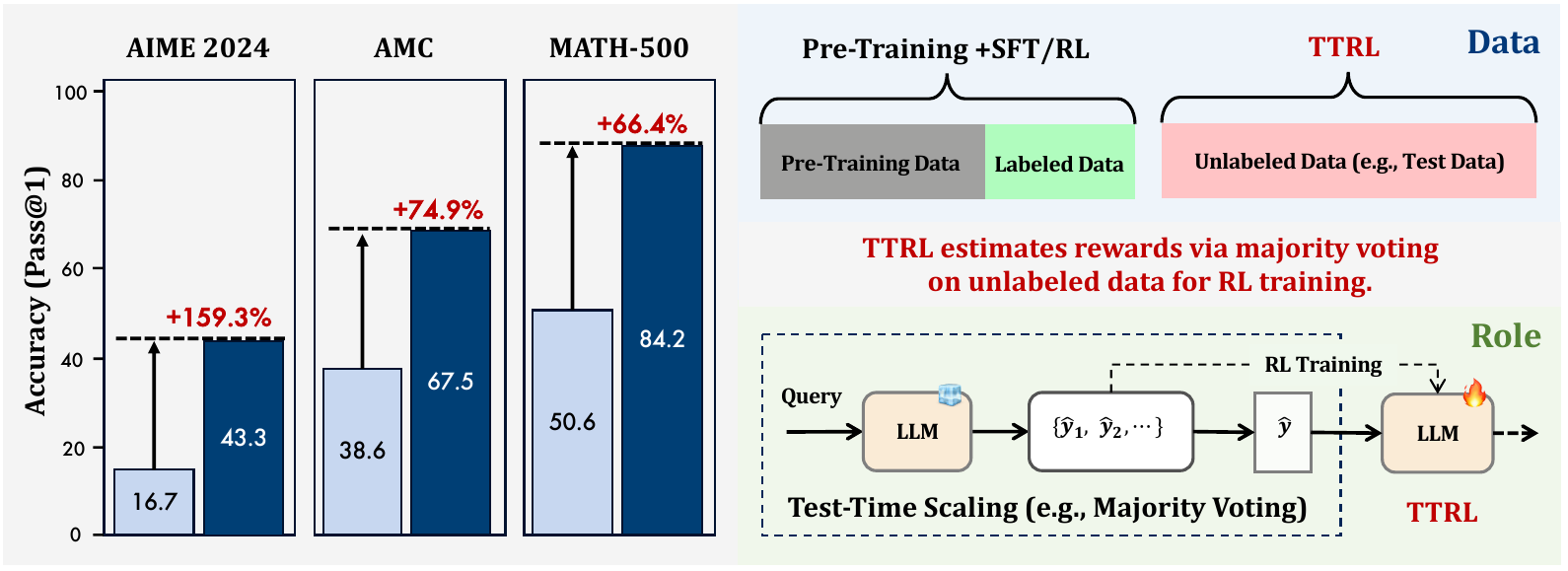}
\caption{
Performance and Position of \method{}.
}
\label{fig:teaser}
\end{figure}

\newpage
\tableofcontents
\newpage

\section{Introduction}\label{sec:intro}

Recent advances in Large Reasoning Models (LRMs), such as DeepSeek-R1~\citep{guo2025deepseek} and OpenAI’s o1~\citep{jaech2024openai}, have demonstrated that Reinforcement Learning (RL) is essential for enhancing long chain-of-thought (CoT) reasoning~\citep{wei2022chain} through training on expensive human-annotated data.
These models achieve remarkable performance on a range of highly challenging tasks. For example, OpenAI’s o3 attains a $75.7\%$ success rate on ARC-AGI-1.
However, complex and unlabeled questions continuously emerge, posing significant challenges.
For instance, o3 solves only $4\%$ of problems on the recently released ARC-AGI-2 benchmark (2025)~\footnote{\url{https://arcprize.org/}}.
Addressing such tasks typically involves scaling up training with more data and computational resources, and it may still fail to yield strong performance on these tasks.
\citet{silver2025welcome} has recently advocated for a transition to the ``era of experience,’’ emphasizing the limitations of existing AI systems that rely heavily on human supervision, as well as the importance of enabling models to self-evolve through experience.

Further building upon the substantial progress of LRMs, it naturally motivates a promising direction in which AI systems autonomously improve via RL on unlabeled data by directly engaging in self-experience and learning, thereby pushing the boundaries of RL and further advancing the frontier of AI capabilities.
Such self-evolvement can be broadly categorized into two modes: adaptation to test-time data, which enables models to tackle harder benchmarks such as ARC-AGI-2, and training on external unlabeled data, which unlocks more training data beyond labeled corpora.
This work focuses on the adaptation to test-time data, which has been extensively studied under the paradigm of Test-Time Training (TTT)~\citep{sun2019test,sun2024learning,behrouz2024titans,akyurek2024surprising}.
TTT has received increasing attention recently. These approaches adapt model parameters at test time by exploiting the structure and distributional properties of incoming test data.

Therefore, we aim to fully advance AI evolution by updating models at test time using RL, thereby enhancing their generalization to previously unseen data. However, this introduces a critical challenge: \textit{How to obtain rewards for RL at test-time?}
This also highlights a broader limitation of current RL approaches.
Despite their promise, most existing methods still rely heavily on labeled data, which significantly limits their scalability.
As real-world tasks continue to increase in both complexity and volume, large-scale annotation for RL becomes increasingly impractical, posing a substantial barrier to the continual improvement of state-of-the-art models.

We introduce Test-Time Reinforcement Learning (\method), which performs test-time training through RL.
\method{} employs repeated sampling strategies in the rollout phase to accurately estimate the label and compute rule-based rewards, thereby enabling RL on unlabeled data.
By incorporating effective majority voting rewards, \method facilitates efficient and stable RL in the absence of ground truth labels.
As previously highlighted, the emergence of more challenging tasks will inevitably lead to larger proportions of unlabeled data.
\method directly addresses the problem of training models via RL without explicit supervision, investigating a model's ability to explore and learn in this challenging yet critical setting.
Essentially, \method enables the model to generate its own experiences, estimate rewards, and improve its performance over time.

In experiments, applying \method to \texttt{Qwen2.5-Math-7B} results in an improvement on AIME 2024 of $\mathbf{211\%}$ ($12.9$ to $40.2$), with an average gain of $\mathbf{76\%}$ across AIME 2024, AMC, MATH-500, and GPQA.
These improvements are achieved through self-evolution without any labeled training data and further generalize to other tasks.
\method not only enhances performance on pass@1 but also improves TTS through majority voting.
Moreover, our preliminary experiments suggest that \method is effective across models of different scales and types and that it can be integrated with existing RL algorithms.
We also found that \method{} exhibits favorable characteristics such as a high-performance ceiling.
These observations highlight its potential to substantially reduce reliance on human annotations, enabling continual learning and scaling RL to large-scale unsupervised training.
Below are several key takeaways:
\begin{tcolorbox}[takeawaysbox]
\begin{enumerate}[leftmargin=1em]
    \item Majority voting provides effective reward estimation for  \method (\textbf{\S~\ref{sec:experiments}}).
    \item \method{} can exceed its training signal and upper limit maj@n, and closely mirrors the performance of direct training on the test data with ground-truth~(\textbf{\S~\ref{sec:how_well}}).
    \item It is possible to achieve efficient and stable RL in an unsupervised manner~(\textbf{\S~\ref{sec:why_work}}).
\end{enumerate}
\end{tcolorbox}

\section{Test-Time Reinforcement Learning (TTRL)}

Unlike traditional RL, where the agent learns from known reward signals, \method{} operates on unlabeled test data. In other words, the model must learn and adapt without access to explicit supervision. Our task is defined as follows:

\begin{tcolorbox}[colback=lightblue!80,breakable,colframe=black]
We study the problem of training a pre-trained model during test time using RL without ground-truth labels.
We call this setting Test-Time Reinforcement Learning.
\end{tcolorbox}

\subsection{Methodology}

\begin{figure}[!h]
  \centering
  \includegraphics[width=1.0\textwidth]{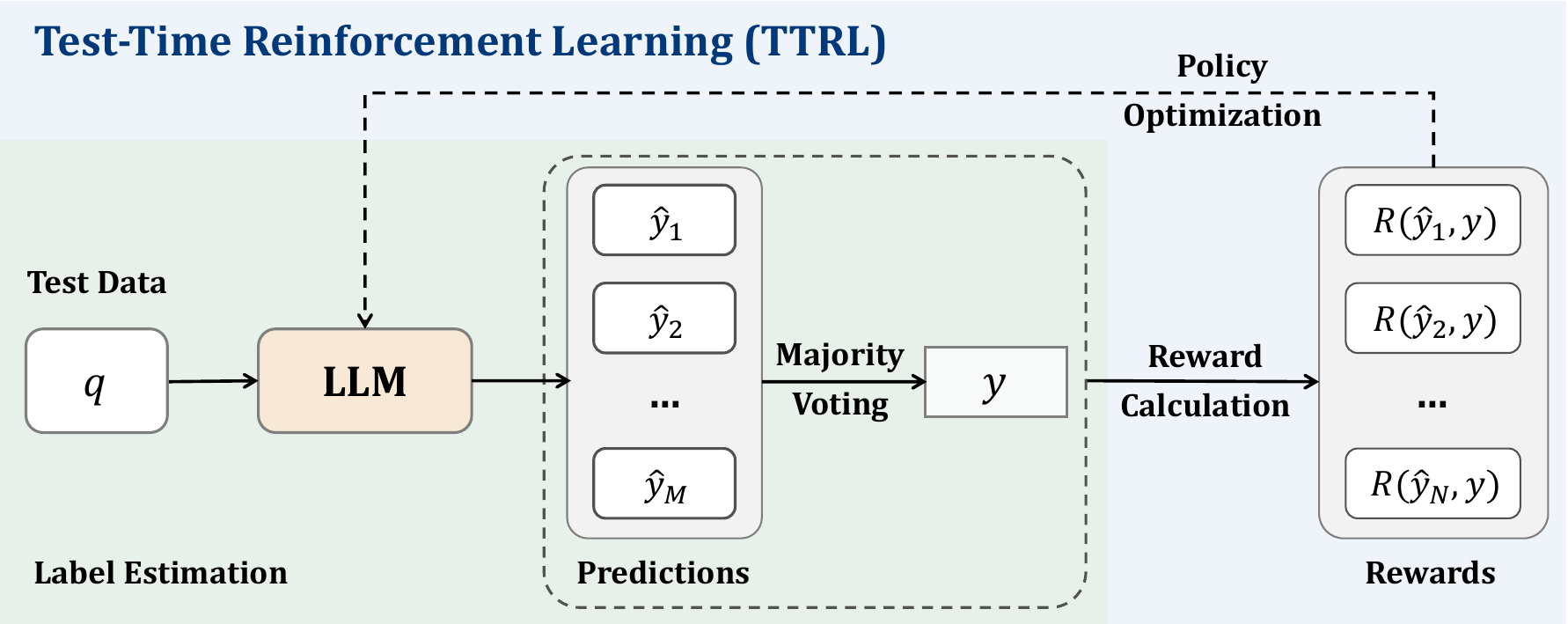}
  \caption{\method{} combines both Test-Time Scaling (TTS) and Test-Time Training (TTT).}
  \label{fig:overview}
\end{figure}

Figure~\ref{fig:overview} illustrates how our approach, \method, tackles this challenge.
Given a state represented by the prompt $x$, the model acts by producing an output $y$ sampled from a policy $\pi_\theta(y \mid x)$ parameterized by $\theta$. To construct a reward signal without ground-truth labels, we generate multiple candidate outputs $\{y_1, y_2, \ldots, y_N\}$ from the model through repeated sampling.
A consensus output $y^*$ is derived, for instance, by majority voting or another aggregation method, serving as a proxy for the optimal action. The environment then provides a reward $r(y, y^*)$ based on the alignment between the sampled action $y$ and the consensus action $y^*$. The RL objective is thus to maximize the expected reward:
\begin{align}
\max_\theta \mathbb{E}_{y \sim \pi_\theta(\cdot \mid x)}[r(y, y^*)],
\end{align}
and parameters $\theta$ are updated through gradient ascent:
\begin{align}
\theta \leftarrow \theta + \eta \nabla_\theta \mathbb{E}_{y \sim \pi_\theta(\cdot \mid x)}[r(y, y^*)],
\end{align}
where $\eta$ denotes the learning rate. This approach enables the model to adapt during inference, effectively improving its performance on distribution-shifted inputs without the need for labeled data.

\newpage

\begin{lstlisting}[
    rulecolor=\color{black},
    label={lst:maj_reward_fn},
    caption={The pseudo-code of the majority voting reward function.},
    abovecaptionskip=2pt,
    belowcaptionskip=7pt,
    language=Python,
]
from collections import Counter

def majority_voting_reward_fn(outputs):
    """
    Assigns a reward of 1 to each output whose extracted answer matches the majority answer, otherwise 0.
    """
    # Extract answers from each output
    answers = [extract_answer(output) for output in outputs]

    # Find the majority answer
    counts = Counter(answers)
    majority_answer, _ = counts.most_common(1)[0]

    # Assign rewards: 1 if matches majority, else 0
    rewards = [1 if ans == majority_answer else 0 for ans in answers]
    return rewards

outputs = llm.generate(problem, n=N)
rewards = majority_voting_reward_fn(outputs)
\end{lstlisting}

\subsection{Majority Voting Reward Function}\label{sec:majority-voting-reward}

The majority voting reward is determined by first estimating a label through majority voting.
This estimated label is then used to calculate rule-based rewards, which serve as the final rewards.
Given a question \( x \), we first input \( x \) into the LLM to generate a set of outputs.
An answer extractor then processes these outputs to obtain the corresponding predicted answers, denoted as \( P = \{ \hat{y}_i \}_{i=1}^{N} \).
We first follow Equation~\ref{eq:tti} over \( P \) to estimate a label, with majority voting as the scoring function \( s(y, x) \) to get \( y \), the most frequently occurring prediction in \( P \).
The majority-voted prediction \( y \) is then used as the estimated label to compute rule-based rewards~\citep{guo2025deepseek}. The reward function is:
\begin{align}
R(\hat{y}_i, y) =
\begin{cases}
1, & \text{if } \hat{y}_i = y, \\
0, & \text{otherwise}.
\end{cases}
\end{align}
Listing~\ref{lst:maj_reward_fn} presents the pseudo-code of the reward function.

\section{Experiments}\label{sec:experiments}

\subsection{Experimental Setup}

\paragraph{Models}
To evaluate the generalizability of \method across different backbone models, we conduct experiments using both base and instruct models of various scales. In addition, we carry out experiments on leading LRMs to demonstrate that \method can improve model performance even after costly post-training. The models we experiment with are as follows:
\begin{itemize}[leftmargin=2em]
    \item \textbf{Qwen Family:} Qwen2.5-Math-1.5B~\citep{yang2024qwen2}, Qwen2.5-Math-7B~\citep{yang2024qwen2}, Qwen2.5-7B~\citep{qwen2.5}, Qwen2.5-32B~\citep{qwen2.5}, Qwen3-8B (thinking mode \& non-thinking mode)~\citep{qwen2.5};
    \item \textbf{LLaMA Family:} LLaMA-3.1-8B-Instruct~\citep{grattafiori2024llama}, LLaMA-3.2-3B-Instruct~\citep{grattafiori2024llama}, LLaMA-3.2-3B-Oat-Zero~\citep{liu2025understanding};
    \item \textbf{Mistral Family:} Mistral-Nemo-Instruct-2407~\citep{MistralAI-NeMo}, Ministral-8B-Instruct-2410~\citep{Ministral-8B-Instruct};
    \item \textbf{DeepSeek Family:} DeepSeek-Math-7B-Instruct~\citep{shao2024deepseekmath}, DeepSeek-R1-LLaMA-8B~\citep{guo2025deepseek};
    \item \textbf{Others:} Skywork-OR1-Math-7B~\citep{skywork-or1-2025};
\end{itemize}

\paragraph{Benchmarks}
We evaluate \method on GPQA-Diamond~\citep{rein2024gpqa}, a challenging and high-quality subset of the Graduate-Level Google-Proof Question Answering benchmark, and $3$ mathematical reasoning benchmarks: AIME 2024~\citep{li2024numinamath}, AMC~\citep{li2024numinamath}, and MATH-500~\citep{hendrycks2021measuring}.

\begin{table}[!t]
\centering
\caption{Main results of \method{} on each task.
$^{*}$ indicates that Qwen3-8B is evaluated in non-thinking mode within a 3k context. Figure~\ref{fig:lrm-results} provides results within a 32k context.
}
\label{tab:main}
\resizebox{.8\linewidth}{!}{
\begin{tabular}{l|cccc|c}
\toprule
\textbf{Name} & \textbf{AIME 2024} & \textbf{AMC} & \textbf{MATH-500} & \textbf{GPQA} & \textbf{Avg} \\
\midrule
\multicolumn{6}{c}{\textbf{Math Base Models}} \\
\midrule
Qwen2.5-Math-1.5B & $7.7$ & $28.6$ & $32.7$ & $24.9$ & $23.5$ \\
\rowcolor{lightblue!100}w/ \method{} & $15.8$ & $48.9$ & $73.0$ & $26.1$ & $41.0$ \\
\rowcolor{lightblue!100}$\Delta$ & \textcolor{red}{$+8.1$} & \textcolor{red}{$+20.3$} & \textcolor{red}{$+40.3$} & \textcolor{red}{$+1.2$} & \textcolor{red}{$+17.5$} \\
& $\uparrow105.2\%$ & $\uparrow71.0\%$ & $\uparrow123.2\%$ & $\uparrow4.8\%$ & $\uparrow74.4\%$ \\
\midrule
Qwen2.5-Math-7B & $12.9$ & $35.6$ & $46.7$ & $29.1$ & $31.1$ \\
\rowcolor{lightblue!100}w/ \method{} & $40.2$ & $68.1$ & $83.4$ & $27.7$ & $54.9$ \\
\rowcolor{lightblue!100}$\Delta$ & \textcolor{red}{$+27.3$} & \textcolor{red}{$+32.5$} & \textcolor{red}{$+36.7$} & \textcolor{darkgreen}{$-1.4$} & \textcolor{red}{$+23.8$} \\
& $\uparrow211.6\%$ & $\uparrow91.3\%$ & $\uparrow78.6\%$ & $\downarrow4.8\%$ & $\uparrow76.5\%$ \\
\midrule
\multicolumn{6}{c}{\textbf{Vanilla Base Models}} \\
\midrule
Qwen2.5-7B & $7.9$ & $34.8$ & $60.5$ & $31.8$ & $33.8$ \\
\rowcolor{lightblue!100}w/ \method{} & $23.3$ & $56.6$ & $80.5$ & $33.6$ & $48.5$ \\
\rowcolor{lightblue!100}$\Delta$ & \textcolor{red}{$+15.4$} & \textcolor{red}{$+21.8$} & \textcolor{red}{$+20.0$} & \textcolor{red}{$+1.8$} & \textcolor{red}{$+14.7$} \\
& $\uparrow194.9\%$ & $\uparrow62.6\%$ & $\uparrow33.1\%$ & $\uparrow5.7\%$ & $\uparrow43.7\%$ \\
\midrule
Qwen2.5-32B & $7.9$ & $32.6$ & $55.8$ & $33.2$ & $32.4$ \\
\rowcolor{lightblue!100}w/ \method{} & $24.0$ & $59.3$ & $83.2$ & $37.7$ & $51.1$ \\
\rowcolor{lightblue!100}$\Delta$ & \textcolor{red}{$+16.1$} & \textcolor{red}{$+26.7$} & \textcolor{red}{$+27.4$} & \textcolor{red}{$+4.5$} & \textcolor{red}{$+18.7$} \\
& $\uparrow203.8\%$ & $\uparrow81.9\%$ & $\uparrow49.1\%$ & $\uparrow13.6\%$ & $\uparrow57.7\%$ \\
\midrule
\multicolumn{6}{c}{\textbf{Instruct Models}} \\
\midrule
LLaMA3.1-8B & $4.6$ & $23.3$ & $48.6$ & $30.8$ & $26.8$ \\
\rowcolor{lightblue!100}w/ \method{} & $10.0$ & $32.3$ & $63.7$ & $34.1$ & $35.0$ \\
\rowcolor{lightblue!100}$\Delta$ & \textcolor{red}{$+5.4$} & \textcolor{red}{$+9.0$} & \textcolor{red}{$+15.1$} & \textcolor{red}{$+3.3$} & \textcolor{red}{$+8.2$} \\
& $\uparrow117.4\%$ & $\uparrow38.6\%$ & $\uparrow31.1\%$ & $\uparrow10.7\%$ & $\uparrow30.6\%$ \\
\midrule
Qwen3-8B$^{*}$ & $26.9$ & $57.8$ & $82.3$ & $48.1$ & $53.8$ \\
\rowcolor{lightblue!100}w/ \method{} & $46.7$ & $69.1$ & $89.3$ & $53.0$ & $64.5$ \\
\rowcolor{lightblue!100}$\Delta$ & \textcolor{red}{$+19.8$} & \textcolor{red}{$+11.3$} & \textcolor{red}{$+7.0$} & \textcolor{red}{$+4.9$} & \textcolor{red}{$+10.8$} \\
& $\uparrow73.6\%$ & $\uparrow19.6\%$ & $\uparrow8.5\%$ & $\uparrow10.2\%$ & $\uparrow20.0\%$ \\
\bottomrule
\end{tabular}
}
\vspace{-4mm}
\end{table}

\paragraph{Evaluation Setup}

We apply \method to each benchmark individually and then evaluate.
We set the maximum generation length to $3072$ tokens, unless otherwise specified.
For the \textbf{main experiments}, following DeepSeek-R1~\citep{guo2025deepseek}, we adopt the pass@$k$ evaluation protocol~\citep{chen2021evaluating} and report pass@1 using non-zero temperature sampling. Specifically, we generate $16$ responses ($4$ for 32k context) per question using a temperature of $0.6$ and a top-$p$ value of $0.95$. The pass@1 score is computed as:
\[
\text{pass@1} = \frac{1}{k} \sum_{i=1}^{k} p_i,
\]
where $p_i$ indicates whether the $i$-th response is correct. 
For the \textbf{analysis and additional experiments on \texttt{Qwen2.5-MATH}}, we evaluate using greedy decoding to report pass@1, to ensure a fair comparison with previous works.
Appendix~\ref{appx:training_metrics} presents a set of training-time metrics we used to monitor the performance of \method and analyze its training dynamics in the absence of ground-truth labels.

\paragraph{Baselines}
Since the use of TTT for reasoning has not been previously explored, we primarily compare it with the backbone model to validate whether \method can achieve effective improvements through self-evolution.
Appendix~\ref{appx:additional-results} presents additional experimental results comparing \method with previous state-of-the-art RL approaches for reasoning.

\paragraph{Implementation Details}
We independently apply GRPO~\citep{shao2024deepseekmath} on each benchmark to implement \method{}.
For hyperparameters, we use a cosine learning rate schedule with a peak value of $5 \times 10^{-7}$ and adopt the AdamW optimizer for the policy model.
For rollout, we sample $64$ responses using a temperature of $0.6$ ($1.0$ for \texttt{Qwen2.5-Math} and LRMs) for voting-based label estimation and downsample $32$ responses per prompt for training.
Evidence shows that our vote-then-sample strategy effectively reduces computational costs while still achieving strong performance.
The maximum generation length is set to $32{,}768$ tokens for LRMs and $3{,}072$ tokens for all other models.
We set the number of episodes to $10$, $30$, and $80$ for MATH-500, AMC, and AIME 2024, respectively, based on the dataset size.
All experiments were conducted on 8 * NVIDIA A100 80GB GPUs.

\subsection{Main Results}\label{sec:main_results}

\paragraph{TTRL performs well on most tasks and models.}

\begin{wraptable}{r}{0.57\textwidth}
\vspace{-4mm}
\centering
\caption{Performance of \method on various models.}
\label{tab:main-1}
\resizebox{0.98\linewidth}{!}{
\begin{tabular}{l|ccc}
\toprule
\textbf{Name} & \textbf{AIME} & \textbf{AMC} & \textbf{MATH-500} \\
\midrule
\multicolumn{4}{c}{\textbf{LLaMA Family}} \\
\midrule
LLaMA-3.2-3B-Oat-Zero & 0.8 & 15.1 & 41.9 \\
\rowcolor{lightblue!100}w/ \method & 3.3 & 25.3 & 55.7 \\
\rowcolor{lightblue!100}$\Delta$ & \textcolor{red}{$+2.5$} & \textcolor{red}{$+10.2$} & \textcolor{red}{$+13.8$} \\
\midrule
LLaMA-3.2-3B-Instruct & 6.0 & 19.4 & 43.9 \\
\rowcolor{lightblue!100}w/ \method & 13.3 & 31.3 & 61.6 \\
\rowcolor{lightblue!100}$\Delta$ & \textcolor{red}{$+7.3$} & \textcolor{red}{$+11.9$} & \textcolor{red}{$+17.7$} \\
\midrule
\multicolumn{4}{c}{\textbf{Mistral Family}} \\
\midrule
Mistral-Nemo-Instruct & 0.8 & 15.4 & 40.8 \\
\rowcolor{lightblue!100}w/ \method & 0 & 24.8 & 51.0 \\
\rowcolor{lightblue!100}$\Delta$ & \textcolor{darkgreen}{$-0.8$} & \textcolor{red}{$+9.4$} & \textcolor{red}{$+10.2$} \\
\midrule
Ministral-8B-Instruct & 1.3 & 19.7 & 52.4 \\
\rowcolor{lightblue!100}w/ \method & 3.3 & 28.9 & 57.8 \\
\rowcolor{lightblue!100}$\Delta$ & \textcolor{red}{$+2.0$} & \textcolor{red}{$+9.2$} & \textcolor{red}{$+5.4$} \\
\midrule
\multicolumn{4}{c}{\textbf{DeepSeek Family}} \\
\midrule
DeepSeek-Math-7B-Instruct & 1.9 & 16.3 & 42.3 \\
\rowcolor{lightblue!100}w/ \method & 2.5 & 22.9 & 52.4 \\
\rowcolor{lightblue!100}$\Delta$ & \textcolor{red}{$+0.6$} & \textcolor{red}{$+6.6$} & \textcolor{red}{$+10.1$} \\
\midrule
DeepSeek-R1-LLaMA-8B & 51.7 & 81.6 & 89.6 \\
\rowcolor{lightblue!100}w/ \method & 69.2 & 88.9 & 90.9 \\
\rowcolor{lightblue!100}$\Delta$ & \textcolor{red}{$+17.5$} & \textcolor{red}{$+7.3$} & \textcolor{red}{$+1.3$} \\
\bottomrule
\end{tabular}
}
\vspace{-3mm}
\end{wraptable}

Table~\ref{tab:main} presents the main results. 
We apply \method to $6$ models spanning $4$ model families, $2$ model types, and $3$ model sizes, consistently demonstrating substantial improvements across $4$ highly challenging benchmarks.
On the demanding mathematical reasoning benchmark AIME 2024, \method achieves a minimum improvement of $105\%$ across all $6$ models.
Moreover, applying \method to a 1.5B model leads to a significant gain of up to $40.3$ points on the MATH-500.
Recently, \citet{shao2025spurious} demonstrated the importance of evaluating different models for RL-based methods to validate experimental conclusions.
Therefore, we additionally report results on a broader range of models from various model families, such as DeepSeek-R1-LLaMA-8B, an LRM from DeepSeek trained on the LLaMA model.
Table~\ref{tab:main-1} presents the results.
As shown, \method continues to exhibit consistent effectiveness.
Furthermore, as shown in Appendix~\ref{appx:additional-results}, despite relying solely on self-evolution using unlabeled test data, \method achieves performance comparable to existing RL-based models that are trained on large-scale labeled datasets.

\paragraph{TTRL performs well on LRMs.}
\begin{wrapfigure}[14]{r}{0.47\textwidth}
  \vspace{-4mm}
  \centering
  \includegraphics[width=0.47\textwidth]{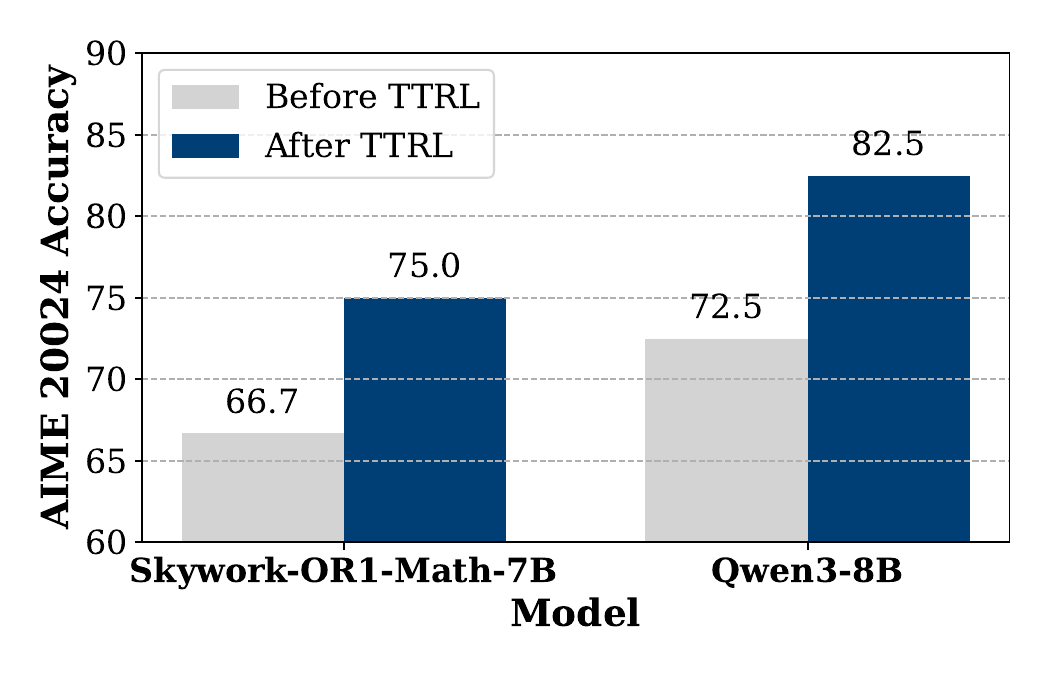}
  \caption{\method\ on LRMs.}
  \label{fig:lrm-results}
\end{wrapfigure}

With the rapid progress in RL and TTS, LRMs are becoming increasingly central.
To further examine whether \method remains effective on LRMs that have undergone expensive post-training, especially on highly challenging tasks, we evaluate two other powerful LRMs.
Figure~\ref{fig:lrm-results} presents the results of applying \method to additional reasoning models. \texttt{Qwen3-8B} is evaluated in thinking mode.
Despite the extensive post-training these models have undergone, \method still achieves substantial performance gains, yielding improvements of approximately 10 points on both backbones.

\paragraph{TTRL naturally scales.} Another noteworthy observation is that as the model size increases (1.5B $\rightarrow$ 7B and 7B $\rightarrow$ 32B), performance consistently improves, highlighting the natural scaling behavior of \method: larger models can produce more accurate majority voting rewards during self-improvement, which leads to more effective learning on new data.

\paragraph{TTRL generalizes well beyond the target task.}

\begin{figure}[ht]
  \centering
  \includegraphics[width=\textwidth]{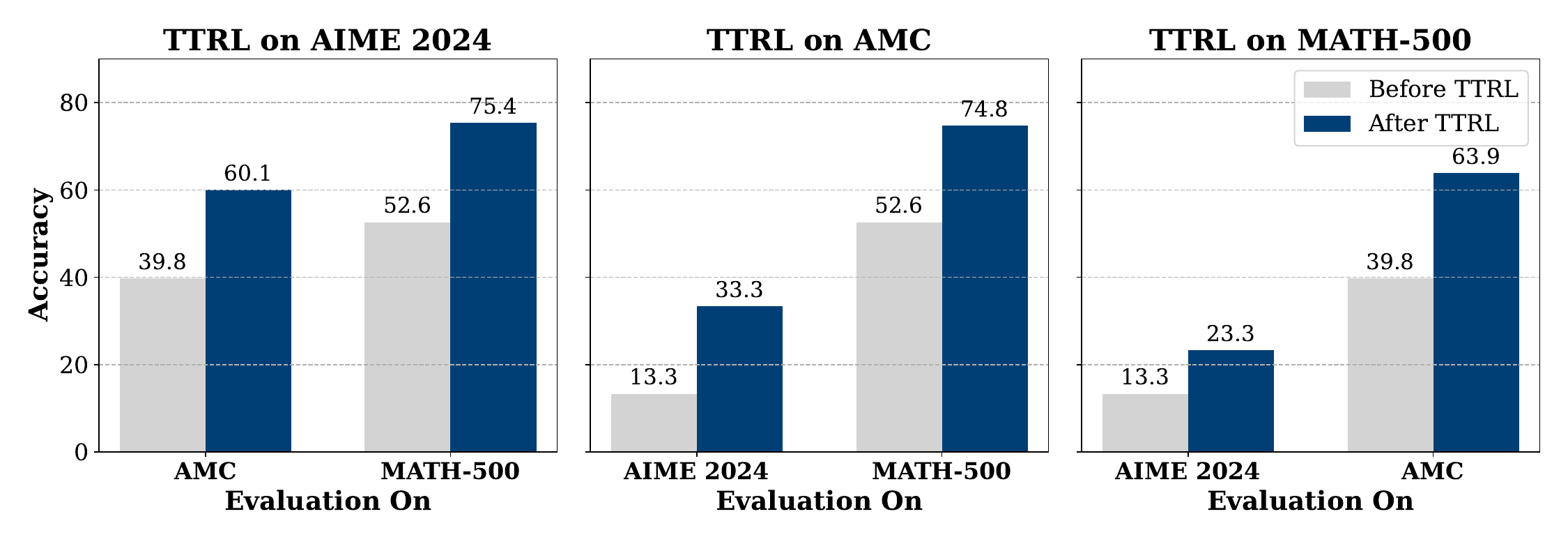}
  \caption{Out-of-distribution performance before and after \method{}.
  }
  \label{fig:ood-ttrl}
\end{figure}

We perform \method on each benchmark and further evaluate pass@1 using greedy decoding on others, with \texttt{Qwen2.5-Math-7B} as the backbone.
Figure~\ref{fig:ood-ttrl} shows the results.
Despite the out-of-distribution nature of this setting, \method achieves substantial improvements across all benchmarks.
This suggests that \method does not rely on overfitting, which would lead to trade-offs on other tasks, but instead acquires generalizable gains during self-improvement.

\paragraph{TTRL is compatible with different RL algorithms.}

\begin{figure}[!ht]
    \centering
    \begin{subfigure}[t]{0.48\textwidth}
        \centering
        \includegraphics[width=\linewidth]{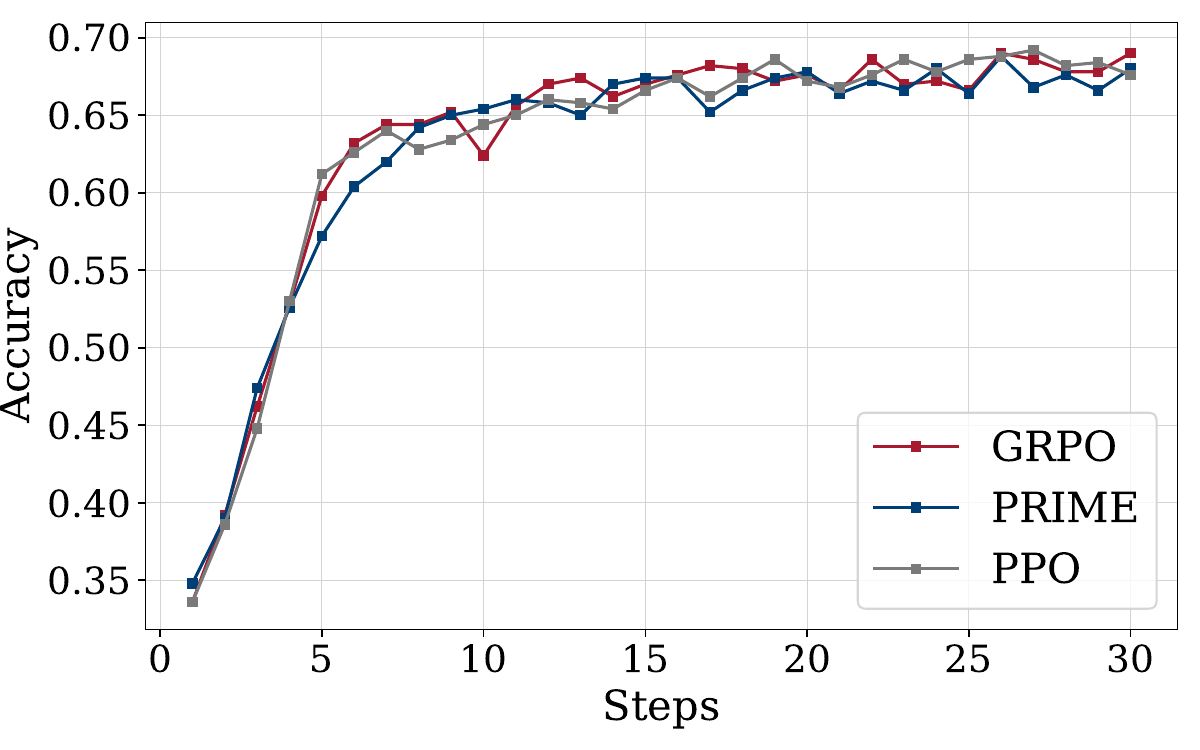}
        \caption{Accuracy Curve.}
        \label{fig:grpo_ppo_acc}
    \end{subfigure}
    \hfill
    \begin{subfigure}[t]{0.48\textwidth}
        \centering
        \includegraphics[width=\linewidth]{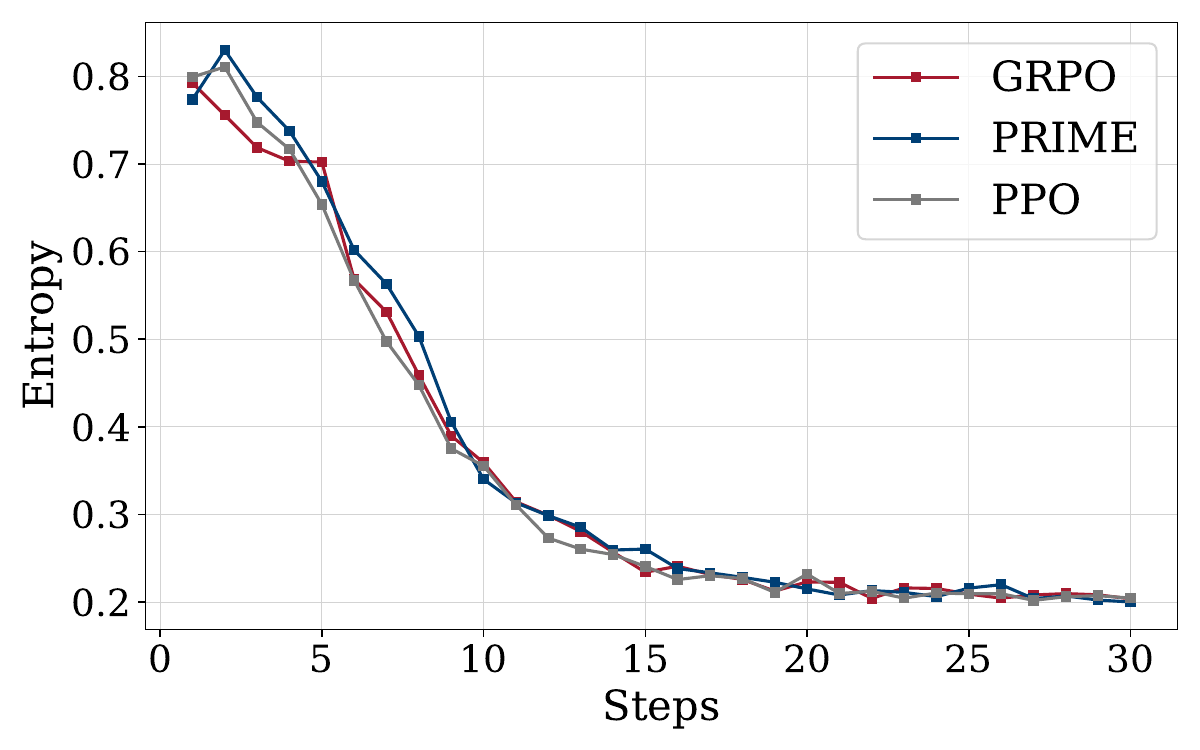}
        \caption{Entropy Curve.}
        \label{fig:grpo_ppo_entropy}
    \end{subfigure}
    \caption{Comparison over steps of different RL algorithms, GRPO, PPO, and PRIME on MATH-500 using \texttt{Qwen2.5-Math-1.5B}.
    }
    \label{fig:grpo_ppo_final}
\end{figure}

We further apply \method using two RL algorithms on MATH-500 to assess its compatibility, which are PPO~\citep{schulman2017proximal}, a value mode based method, and PRIME~\citep{cui2025process}, a process-level RL algorithm.
Figure~\ref{fig:grpo_ppo_final} presents the results.
The performance trajectories of GRPO, PPO, and PRIME are closely aligned.

\paragraph{TTRL achieves sustainable self-evolution through ``online'' and ``RL''.}

\begin{figure}[!ht]
    \centering
    \includegraphics[width=1.0\textwidth]{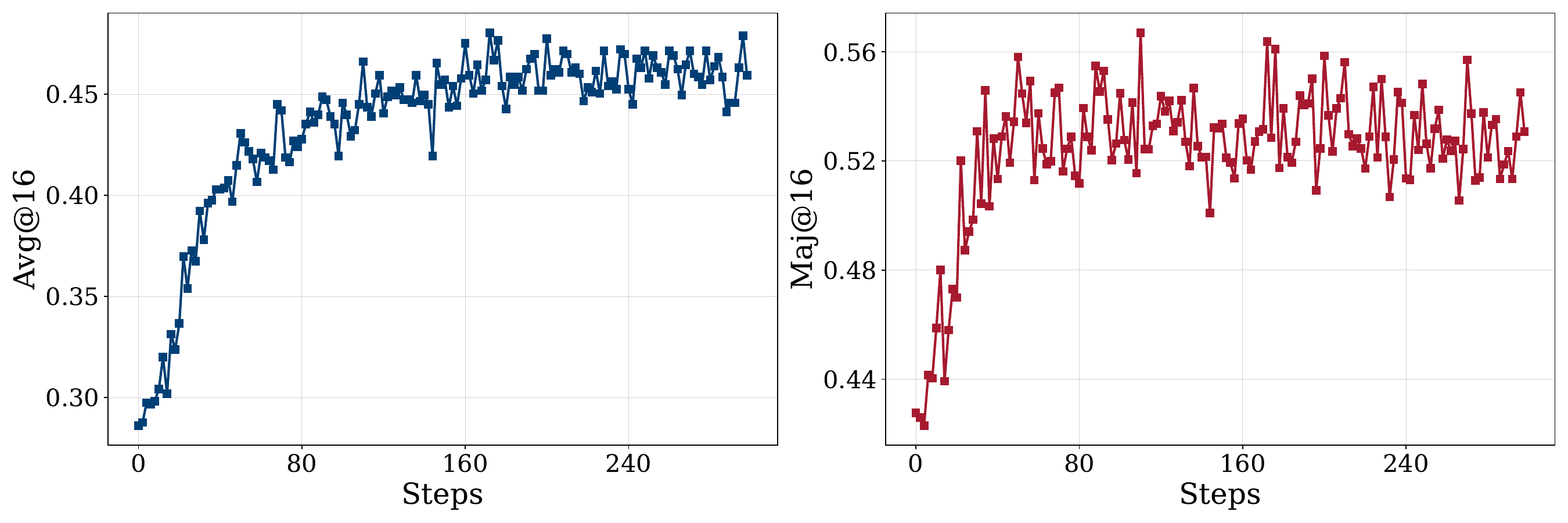}
    \caption{Training dynamics of \method on AMC using \texttt{Qwen2.5-Math-1.5B} as the base model.}
    \label{fig:training_dynamics}
\end{figure}

To gain a deeper understanding of the underlying mechanisms of \method, we conduct an analysis of the model's training dynamics by tracking the average (pass@1/avg@16) and majority (maj@16) scores throughout the training process.
Given that majority voting serves as the basis for generating training signals, examining its performance trajectory is essential for understanding how it functions.
Furthermore, we investigate whether \method improves pass@1 at the cost of a reduction in maj@16 performance.
Figure~\ref{fig:training_dynamics} illustrates the \method training dynamics on AMC with \texttt{Qwen2.5-Math-1.5B} as the base model.
\textbf{It is notable that, as training progresses, both metrics demonstrate a consistent upward trend.}
This indicates that \method is not simply approaching the initial model's majority voting performance.
Due to its dynamic nature, \method can generate higher-quality supervision signals as its capabilities improve.
Moreover, through \method's use of RL for TTT, by converting voting-based pseudo-labels into reward signals, it enhances the effective supervision quality (e.g., accuracy; see Q2~\ref{sec:why_work}), while decoupling learning from the limitations imposed by maj@n.

\section{Analysis and Discussions}\label{sec:discussions}

\subsection{Q1: How Well Can TTRL Perform?}\label{sec:how_well}

\begin{tcolorbox}[takeawaysbox]
\begin{enumerate}[leftmargin=1em]
     \item \method{} surpasses the traditional self-training upper bound, the majority accuracy of the initial model.
     \item The empirical upper bound of \method is direct RL on labeled test data (i.e., training on the test data). \method can approach the performance of this upper bound, highlighting its potential advantages in efficacy over standard \textit{training-evaluation} protocols.
     \item For challenging tasks, \method can reach the empirical upper bound using only a 1.5B model. This demonstrates that LLMs can now efficiently self-evolve through \method, enabling unbounded lifelong learning on large-scale datasets.
\end{enumerate}
\end{tcolorbox}

We analyze the potential performance of \method{} using two upper bounds.
The first upper bound is the maj@n of the initial model.
The second upper bound is direct training on benchmark data, which assumes access to ground-truth labels and thus leaks label information to the policy model.

\begin{figure}[!ht]
  \centering
  \includegraphics[width=\linewidth]{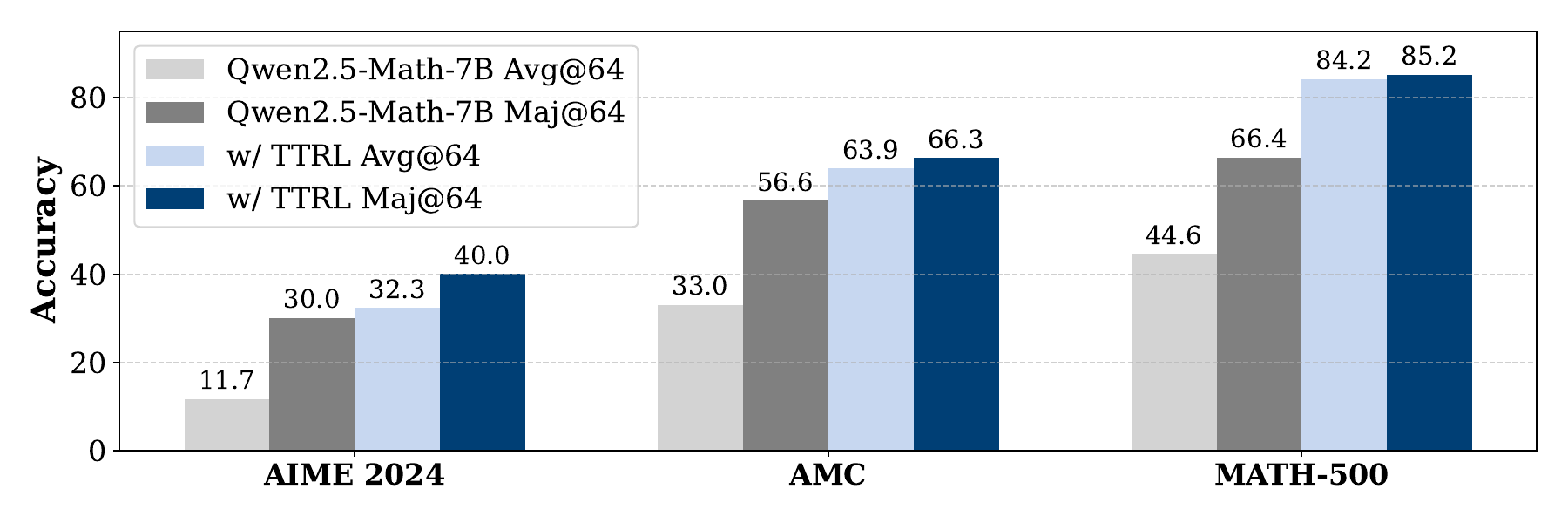}
  \caption{Majority voting performance comparison between the backbone and after \method.
  }
\label{fig:voting_performance_comparison}
\end{figure}

\textbf{TTRL is Supervised by maj@n Yet Surpasses It.}
Since \method utilizes the model’s own majority-voted outputs for RL, this voting-based performance of the initial model can intuitively be regarded as an upper bound of the final performance. This upper bound is also the performance limit of traditional self-training methods~\citep{huang2022large}, which select self-generated CoT through majority voting for supervised fine-tuning (SFT).
However, we observe a surprising phenomenon: after training, the model not only matches but also surpasses the expected upper bound, suggesting that it exceeds the performance limit of the original model, which also serves as its initial supervision signal.
Figure~\ref{fig:training_dynamics} illustrates this remarkable result, where it can be observed that the final avg@16 score exceeds the initial maj@16 score by more than 20 points. 
Furthermore, we perform additional evaluations of \method on \texttt{Qwen2.5-Math-7B} across various benchmarks, using more samples per question to enable more reliable assessment.
Figure~\ref{fig:voting_performance_comparison} shows results.
\textbf{It can be observed that \method avg@64 consistently outperforms \texttt{Qwen2.5-Math-7B} maj@64 across all benchmarks, with a considerable margin.}
Through a self-reinforcing loop, the model \textit{\textbf{``lifts itself up by its own bootstraps''}}, evolving beyond the anticipated performance ceiling.
Moreover, the performance of \method further improves when majority voting is applied.

\begin{wrapfigure}{r}{0.55\textwidth}
    \centering
    \includegraphics[width=\linewidth]{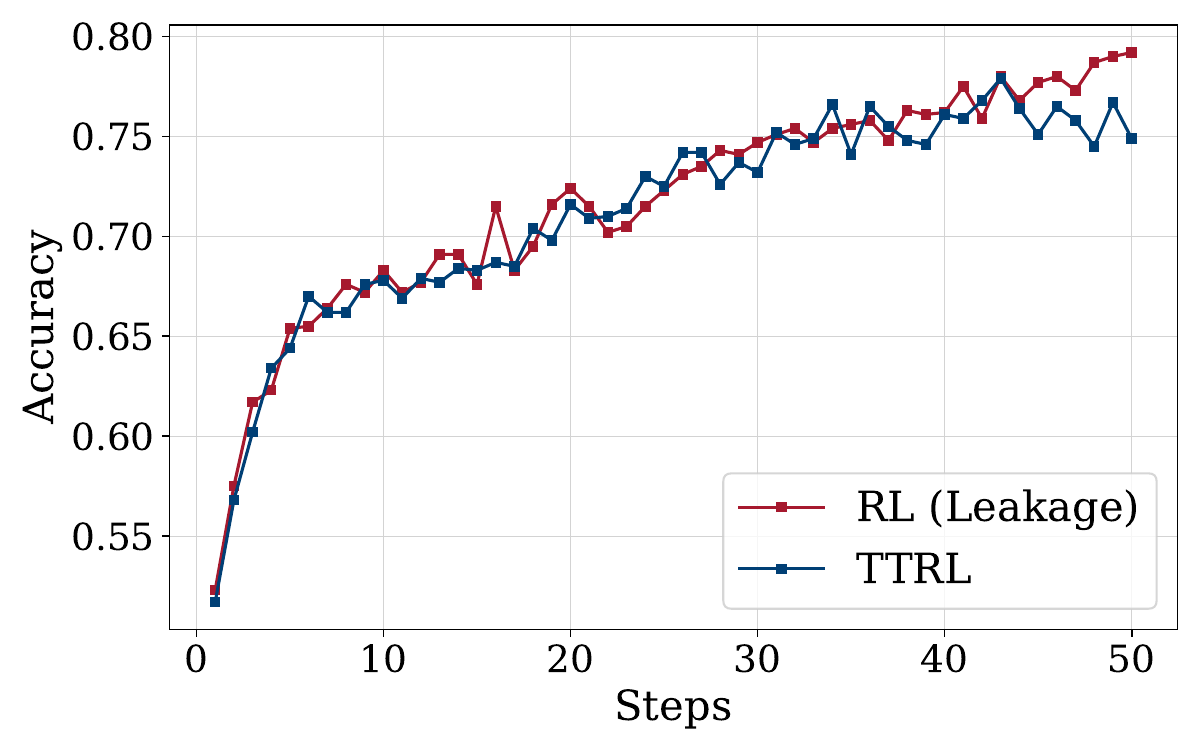}
    \caption{Comparison of RL (Leakage) \textit{vs} \method{}.}
    \label{fig:rl_ttrl}
\end{wrapfigure}

\textbf{TTRL's Performance Gains Approach Training on the Benchmark.}
The motivation of \method is to estimate labels using majority voting to obtain more accurate rewards, facilitating effective self-improvement through RL on the data without ground-truth labels.
Therefore, a natural upper bound of \method is performing RL directly on the test data, denoted as RL (leakage).
Although this setting is rarely adopted or studied due to the issue of information leakage, it represents the most efficient way to improve performance on the particular dataset, with efficiency that far exceeds traditional training-evaluation paradigms.
We use \texttt{Qwen2.5-Math-7B} to perform both \method and RL (leakage) on MATH-500 and conduct evaluations.
Figure~\ref{fig:rl_ttrl} shows results.
Surprisingly, we find that the performance curve of \method closely approaches that of RL (leakage).
This suggests that:

\begin{enumerate}[leftmargin=2em]
    \item \method can achieve a level of self-improvement comparable to that of supervised learning (even in the information leakage scenario) through RL in an unsupervised setting. This indicates its substantial efficiency and performance gains.
    \item \method provides evidence that even small LLMs can now effectively self-improve on input-only challenging tasks through RL, enabling continual learning.
    Results on \texttt{Qwen2.5-Math-1.5B} further support this observation: starting from a subpar performance of $32.7$ on MATH-500, the model improved by $123.2\%$ to reach $73.0$, demonstrating clear self-improvement through \method.
\end{enumerate}

\begin{figure}[!ht]
    \centering
    \includegraphics[width=1.0\textwidth]{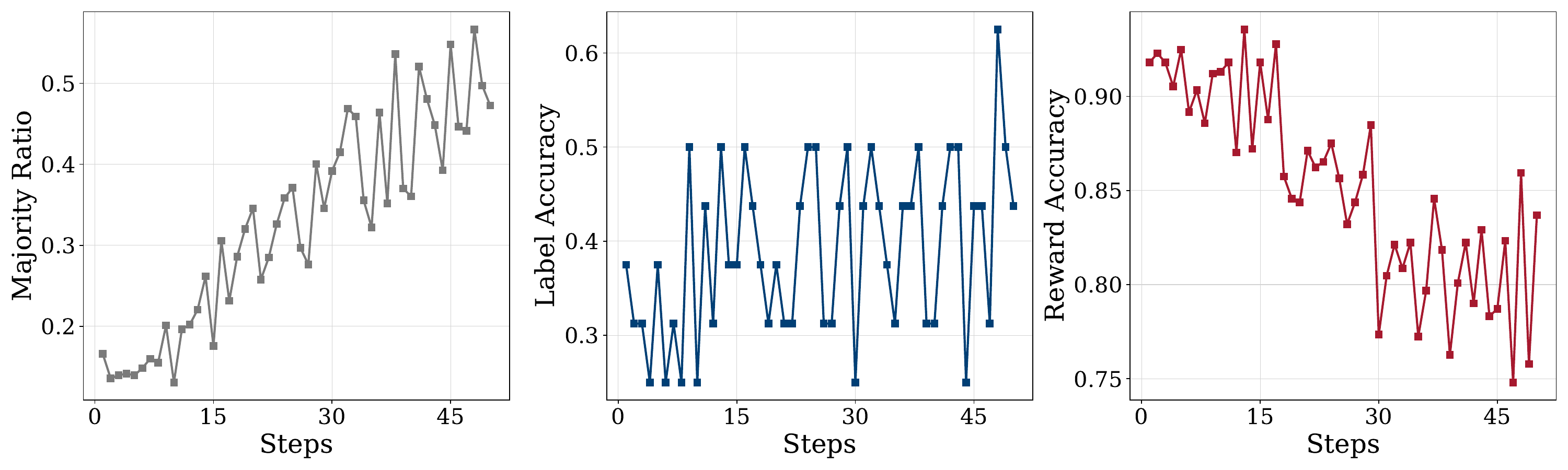}
    \caption{Comparison of Majority Ratio, Label Accuracy, and Reward Accuracy on AIME 2024 over steps. Even with low label accuracy, reward accuracy remains high due to ``\textcolor{darkred}{Lucky Hit}'', allowing \method to provide reliable training signals.}
    \label{fig:lucky-hit}
\end{figure}

\subsection{Q2: Why Does TTRL Work?}\label{sec:why_work}

This section presents a progressive analysis of the factors enabling \method to achieve stable and effective RL under unsupervised conditions.
Our analysis identifies three key factors: label estimation, reward calculation, and online learning.

\paragraph{Label Estimations.}
A direct difference between \method{} and standard RL algorithms is that \method{} involves label estimation, which introduces reward inaccuracies. We believe that \method works despite these inaccuracies due to the following two reasons.
\textbf{(i)} Existing studies have shown that RL can tolerate a certain degree of reward inaccuracy. Moreover, RL tends to generalize better than SFT, which often relies on memorizing training data~\citep{chu2025sft}. In RL, rewards are typically vague and serve primarily as directional signals for exploration, leading to RL's robustness to reward noise~\citep{razin2025makes}. 
\textbf{(ii)} Prior work has also examined what constitutes a good reward model from an optimization perspective, revealing that more accurate reward models are not necessarily better teachers~\citep{wang2020reinforcement}.
Therefore, reward signals estimated by the policy model itself may offer more suitable guidance for learning.

\paragraph{Reward Calculations.}

\begin{wrapfigure}{r}{0.45\textwidth}
    \centering
    \includegraphics[width=0.45\textwidth]{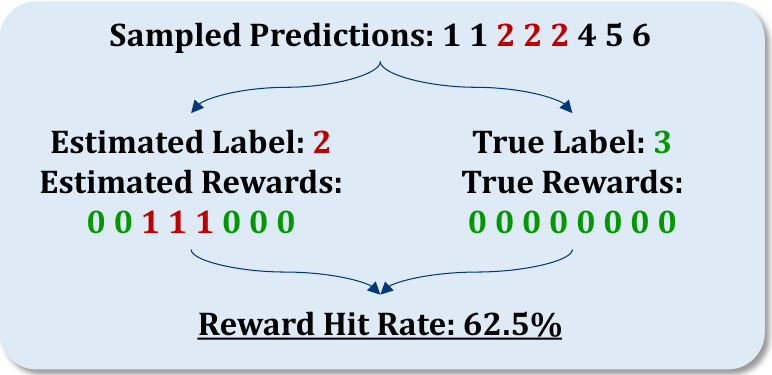}
    \caption{A toy case of ``\textcolor{darkred}{Lucky Hit}''.
    We illustrate a basic numerical prediction scenario to compare reward computation under two conditions: when the model incorrectly estimates the label versus when the ground-truth label is used. As shown on the left, although the estimated label is incorrect, some of the incorrect predictions still differ from the wrong label and therefore receive the correct reward (denoted as \textcolor{darkgreen}{$\mathbf{0}$}).
    }
    \label{fig:reward_calculations}
\end{wrapfigure}

When the model is capable of estimating accurate labels via majority voting, the reward and subsequently training are generally reliable.
However, a natural question arises: \textbf{Why does \method remain effective even when the model fails to estimate accurate labels via majority voting on challenging benchmarks such as AIME 2024?}
The most fundamental reason lies in the mechanism by which the verifier computes rewards in RL.
For tasks such as mathematics, the verifier works based on ``comparison'' to obtain rule-based rewards by checking whether the predicted answer matches the given ``label.''
This mechanism can lead to the phenomenon of ``\textcolor{darkred}{Lucky Hit}'': for an incorrectly predicted answer, even if the estimated label does not match the ground truth label, as long as it differs from the predicted answer, the verifier will still output a negative reward, and this is exactly the correct reward that we expect, as illustrated in Figure~\ref{fig:reward_calculations}.
In other words, it is sufficient that the estimated label differs from the predicted answer for the verifier to assign the correct negative reward.
To provide a more detailed case study, we examine the performance of \method on the AIME 2024 using \texttt{Qwen2.5-Math-7B}.
Figure~\ref{fig:lucky-hit} presents the variation curves of the three metrics, as described in Appendix~\ref{appx:training_metrics}.
We identify two main reasons why \method remains effective on AIME 2024:

\begin{enumerate}[leftmargin=2em]
    \item \textbf{Reward robustness enabled by multiple outputs within a rollout.}
First, rewards are denser than labels, allowing for more opportunities to recover useful reward signals even when the estimated label is inaccurate.
For example, even when the predicted label is incorrect, alternative outputs within the same rollout can still yield correct or high-quality rewards, as shown in Figure~\ref{fig:reward_calculations}, whereas a rollout containing only a single output would not provide such flexibility.
This makes the overall reward signal more robust to errors in pseudo-label estimation.

    \item \textbf{High reward accuracy due to scattered incorrect predictions.}
Second, counterintuitively, when the model has weaker capability, the majority voting rewards of \method may be more accurate.
As shown in Figure~\ref{fig:lucky-hit}, although the initial label estimation through majority voting achieves an accuracy of only $37\%$, the reward accuracy reaches an impressive $92\%$.
By examining the model outputs, we find that this is because the model’s responses are highly scattered and consistently incorrect, as shown in Figure~\ref{fig:reward_calculations}.
A result consistent with this observation is that, for the base model, the most frequently predicted answer accounts for only $16.6\%$ of all predictions, indicating that the outputs are highly scattered.
Therefore, even when the labels are not accurately estimated, due to ``\textcolor{darkred}{Lucky Hit}'', most outputs can still receive correct rewards.
Moreover, the poorer the model’s performance, the more mistakes it tends to make, which paradoxically leads to more accurate reward estimation.
An empirical observation supporting this view is the comparison between the label accuracy and reward accuracy, as shown in Figure~\ref{fig:lucky-hit}.
Although the label accuracy rarely exceeds $50\%$, the reward accuracy remains consistently high, staying above $75\%$.
This high reward accuracy provides a reliable foundation for effective self-improvement on test data.

\end{enumerate}

\paragraph{Online Learning.}

\method is designed based on an online RL approach, whereas traditional self-training and test-time training methods operate in an offline manner. The online nature of \method enables the model to improve its capabilities during the application, which in turn leads to more accurate labels generated through voting. As a result, the quality of the supervision signal improves, allowing for truly sustainable self-evolution. As shown in Figure~\ref{fig:training_dynamics}, this dynamic learning process leads to a complementary improvement of performance in both pass@1 and maj@n.

\subsection{Q3: When Might TTRL Fail?}\label{sec:when_fail}

At the algorithmic level, \method is not fundamentally different from existing RL algorithms and therefore inherits several of their characteristics, such as sensitivity to data difficulty, strong reliance on priors, and risk of collapse under certain conditions.
At the implementation level, these issues are further amplified by the constraints of \method, which estimates labels via majority voting and operates exclusively on test data that is both sparse and previously unseen, potentially resulting in failures in certain scenarios.
In our preliminary experiments, we identified two potential issues:

\paragraph{Inappropriate RL Hyperparameters.}

\begin{wrapfigure}{r}{0.5\textwidth}
    \vspace{-4mm}
    \centering
    \includegraphics[width=0.48\textwidth]{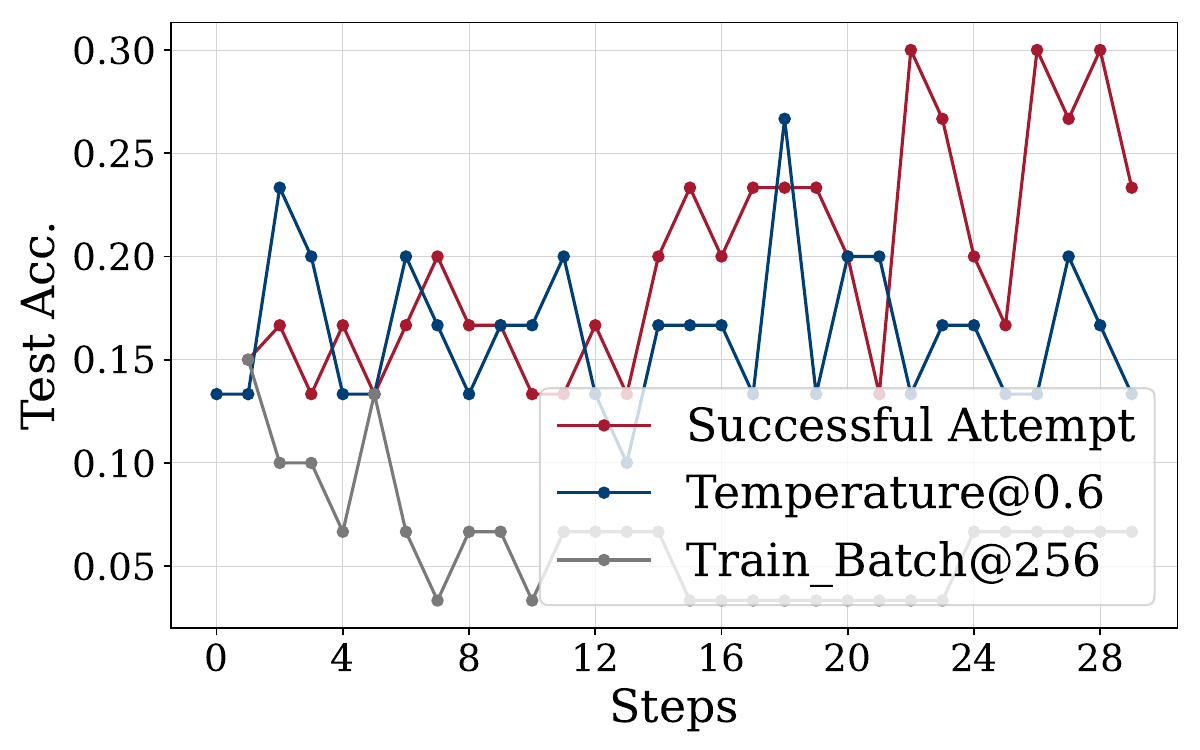}
    \caption{Failed attempts.
    We compare the curves under settings with appropriate parameters versus those with suboptimal temperature and training batch size.
    }
    \label{fig:failed_attempts}
\end{wrapfigure}

Hyperparameter settings play a crucial role in RL training, varying across projects~\footnote{\url{https://github.com/TsinghuaC3I/Awesome-RL-Reasoning-Recipes}} and often leading to training failures.
The influence of hyperparameters is further amplified in \method due to potential noise in reward estimation and the characteristics of the test data.
Figure~\ref{fig:failed_attempts} presents a comparison of several unsuccessful attempts on AIME 2024.
Both of these failed attempts exhibit persistently high entropy that does not diminish throughout training, consistent with findings of prior work~\citep{skywork-or1-2025}.
In our preliminary experiments, we identified two key hyperparameters that can critically affect training stability and success:

\begin{itemize}[leftmargin=1em]
    \item \textbf{Temperature:} Setting the temperature to $1.0$, as opposed to $0.6$, increases the model’s output entropy. This promotes more extensive exploration and allows the model to make better use of its prior knowledge for self-improvement, which is particularly important when addressing challenging benchmarks.
    \item \textbf{Episodes:} Given the substantial variation in size and difficulty across datasets, smaller and more difficult datasets need more episodes to achieve sufficient exploration.
\end{itemize}

\paragraph{Lack of Prior Knowledge on Target Task.}

Prior knowledge plays a crucial role in RL, often determining the success or failure of the \method learning process\footnote{\url{https://ysymyth.github.io/The-Second-Half/}}.
This is mainly because the test data generally exhibits higher difficulty and introduces new features, but \method does not incorporate mechanisms such as data filtering to support curriculum learning.

Therefore, for the same backbone, \method fails if the model’s prior knowledge is insufficient to handle the complexity of the data.
To further validate this hypothesis, we conduct an ablation study on MATH-500.
We divide MATH-500 into five subsets according to its annotated difficulty levels, ranging from $1$ to $5$, and apply \method to each subset independently, using \texttt{Qwen2.5-Math-1.5B}. We then compare the results to those of the backbone, as shown in Table~\ref{tab:math-l1-5}.
We observe that as the question difficulty increases, both the performance improvement and length reduction ratios tend to decrease.
\textbf{This suggests that the available prior knowledge of the backbone is insufficient to support learning on more challenging questions.}

\begin{table}[!t]
\centering
\caption{Performance of \method across the five difficulty levels of MATH-500.}
\label{tab:math-l1-5}
\resizebox{\linewidth}{!}{
\begin{tabular}{@{}llccccc@{}}
\toprule
Metric                   & Name                      & MATH-500-L1 & MATH-500-L2 & MATH-500-L3 & MATH-500-L4 & MATH-500-L5 \\ \midrule
Accuracy     & Backbone                      & $25.9$       & $33.0$       & $36.3$       & $32.5$       & $22.3$       \\
                         & w/ \method                      & $71.2$       & $76.2$       & $76.3$       & $58.7$       & $39.2$       \\ \cmidrule(l){2-7} 
                         & \multicolumn{1}{r}{$\Delta$} & $+45.4$       & $+43.2$       & $+40.0$       & $+26.2$       & $+16.8$       \\
                         &                           & $\uparrow175.3\%$    & $\uparrow130.8\%$    & $\uparrow110.2\%$    & $\uparrow80.4\%$     & $\uparrow75.3\%$     \\ \midrule
Response Len. & Backbone                      & $2{,}339.2$    & $2{,}125.1$    & $2{,}120.6$    & $1{,}775.1$    & $1{,}751.3$    \\
                         & w/ \method                      & $624.3$      & $614.4$      & $672.3$      & $783.5$      & $985.3$      \\ \cmidrule(l){2-7} 
                         & \multicolumn{1}{r}{$\Delta$} & $-1{,}715.0$    & $-1{,}510.6$    & $-1{,}448.3$    & $-991.6$      & $-766.0$      \\
                         &                           & $\downarrow73.3\%$     & $\downarrow71.1\%$     & $\downarrow68.3\%$     & $\downarrow55.9\%$     & $\downarrow43.7\%$     \\ \bottomrule
\end{tabular}
}
\end{table}

\section{Related Works}

\subsection{Test-Time Scaling}

Test-Time Scaling (TTS) is designed to enhance the capabilities of Large Language Models (LLMs) in handling complex tasks by increasing computational resources at test time.
Prior research~\citep{snell2024scaling,liu2025can} indicates that TTS is more efficient than scaling during pre-training~\citep{kaplan2020scaling}. Therefore, reallocating the same computational resources from pre-training to test-time could yield greater improvements in model performance.
Current studies on TTS fall into two categories~\citep{welleck2024decoding}: parallel generation and sequential generation.
Parallel generation involves LLMs producing multiple candidate responses (self-consistency~\citep{wang2022self,chen2023universal}, best-of-N~\citep{stiennon2020learning,nakano2021webgpt}), decision steps (Monte Carlo Tree Search~\citep{zhou2023language,xie2024monte}), or tokens (Reward-guided Search~\citep{deng2023reward, khanov2024args}) during inference.
Subsequently, an aggregation strategy is applied to integrate these candidates, commonly using process reward models~\citep{lightman2023let,wang2023math,zhangopenprm}.
Concurrently, sequential generation focuses on extending the LLMs' output to include longer responses with reflective and chain-of-thought (CoT) processes~\citep{wei2022chain,madaan2023self}.
Although prompting techniques are widely adopted, they are often constrained by the capabilities of the underlying models.
Notably, DeepSeek-R1~\citep{guo2025deepseek} is a representative advancement in this area, achieving extended reasoning capabilities in pre-trained language models through outcome-based reinforcement learning (RL), more specifically group relative policy optimization~\citep{shao2024deepseekmath}.
Compared to the first approach, which requires intensive process-level supervision~\citep{yuan2024free}, the second approach is more scalable due to its reliance on rule-based rewards.

Beyond the aforementioned methods that focus on scaling test-time inference computation, another approach to increasing test-time computing is \textbf{Test-Time Training (TTT)}. We introduce the relationship between these terminologies in Appendix~\ref{sec:appendix_terminology}. While prior work has primarily focused on applications such as video generation and understanding~\citep{hardt2023test,dalal2025one}, and to some extent on large language models~\citep{wang2025test,akyurek2024surprising}, the integration of test-time scaling with reinforcement learning remains largely underexplored.

\subsection{RL for Reasoning}
Reinforcement Learning (RL)~\citep{sutton1998reinforcement} plays a critical role in enhancing the instruction-following capabilities of Large Language Models (LLMs), particularly through approaches like Reinforcement Learning from Human Feedback (RLHF)~\citep{ouyang2022training}.
RLHF aligns base models with human preferences using algorithms such as Proximal Policy Optimization (PPO)~\citep{schulman2017proximal}, where preference modeling is essential.
Recently, Large Reasoning Models (LRMs), such as DeepSeek-R1~\citep{guo2025deepseek}, have demonstrated the significance of RL in improving reasoning abilities using rule-based rewards, as exemplified by GRPO~\citep{shao2024deepseekmath}. Unlike RLHF, which is tailored to open-domain instructions, GRPO is specifically designed to elicit long CoT~\citep{wei2022chain} reasoning in mathematical problem-solving. Recent studies have focused primarily on improving the training stability of rule-based RL methods like GRPO and PPO~\citep{cui2025process,yu2025dapo,liu2025understanding}.
However, these methods typically train LLMs only on supervised training data, while inference involves generating extended CoT reasoning on unseen test problems. Moreover, current RL approaches~\citep{hu2025open,wei2025swe} depend on verifiable outputs—such as solutions in mathematics or code—that can provide reliable reward signals.

Previous studies have explored self-rewarding~\citep{yuan2025selfrewardinglanguagemodels,prasad2024self} and self-play training~\citep{chen2024self} for unlabeled data. However, these works primarily focus on open-domain instruction following~\citep{yuan2025selfrewardinglanguagemodels,chen2024self} rather than mathematical reasoning or employ preference-based optimization strategies~\citep{prasad2024self} such as DPO~\citep{rafailov2023direct} instead of online reinforcement learning algorithms.
In addition to these studies, we identified several concurrent works~\citep{xu2025genius,zhang2025right,zhao2025absolutezeroreinforcedselfplay}, that explore self-supervised and semi-supervised reasoning using reinforcement-like methods. The key distinction lies in reward estimation: we employ majority voting, which is derived from the model itself and mitigates reward hacking.
Recently, \cite{wang2025reinforcement} demonstrated that using a single training example to incentivize the mathematical reasoning capabilities of LLMs is effective, showing substantial improvements even under minimal supervision.
We acknowledge that future research integrating the insights and strengths of these approaches could lead to more robust reasoning models in the era of experience~\citep{silver2025welcome}.
\textbf{\method{} offers a preliminary attempt at RL with self-labeled rewards, advancing toward learning from streams of experience.}

\section{Conclusion}

In this paper, we propose Test-Time Reinforcement Learning (\method{}), a novel framework for training large language models with Reinforcement Learning (RL) on test data without access to ground-truth labels.
A key component of \method{} is its majority voting reward function, which generates rule-based rewards based on consensus among model predictions.
Our experiments demonstrate the strong potential of \method{}, achieving consistent improvements across a variety of models and tasks.
We view \method{} as a preliminary step toward RL with self-labeled rewards, marking an important direction of learning from continuous streams of experience.

\section{Limitations and Future Works}\label{sec:limitations_and_future_works}

\paragraph{Limitations}
This work represents an initial exploration of test-time reinforcement learning using self-labeled rewards.
While our experimental results are promising, several aspects require further investigation.
In particular, we plan to conduct a more in-depth analysis of the impact of prior knowledge and hyperparameter configurations, both of which play critical roles in reinforcement learning dynamics. We will provide comprehensive discussions and ablation studies in future revisions of this paper.

\paragraph{Future Works}
Building on our findings, we identify several directions for future research:

\begin{itemize}[leftmargin=1em]
    \item \textbf{Theoretical Analysis:} Developing a formal convergence analysis of \method{}, particularly focusing on its ability to optimize toward the two upper bounds in $\S$~\ref{sec:how_well}.
    \item \textbf{Online Learning with Streaming Data:} Extending \method{} to real-time learning scenarios, where models interact with continuously arriving data and adapt dynamically, that is Test-Time Adaptation~\citep{liang2025comprehensive}.
    \item \textbf{Large-Scale Self-Supervised RL Training:} Scaling up \method{} to massive datasets and models to explore its potential in self-supervised regimes without human-labeled data.
    \item \textbf{Agentic Tasks and Scientific Discovery:} Applying \method{} to more complex, open-ended domains such as agentic tasks and multi-step scientific reasoning.
\end{itemize}

\newpage

\bibliography{colm2025_conference}
\bibliographystyle{colm2025_conference}

\newpage

\appendix

\section{Additional Results}\label{appx:additional-results}

Table~\ref{tab:additional-results} shows pass@1 results using greedy decoding.
For the two base models, we further include comparisons with their instruct versions that have undergone large-scale post-training.
In addition, we include for reference current leading ``R1-Zero-Like'' models with similar backbones, which are extensively trained using RL:
DeepSeek-R1-Distill-1.5B\&7B~\citep{guo2025deepseek}, SimpleRL-Zero-7B~\citep{zeng2025simplerlzooinvestigatingtamingzero}, PRIME-Zero-7B~\citep{cui2025process}, OpenReasoner-Zero-7B~\citep{hu2025openreasonerzeroopensourceapproach}, Oat-Zero-1.5B\&7B~\citep{liu2025understanding}, and LIMR~\citep{li2025limr}.
\textbf{Note that \method{} has a different setup from the previous models, which makes the comparison seem unfair.}

\begin{figure}[!ht]
    \centering
    \begin{subfigure}[t]{0.48\textwidth}
        \centering
        \includegraphics[width=\linewidth]{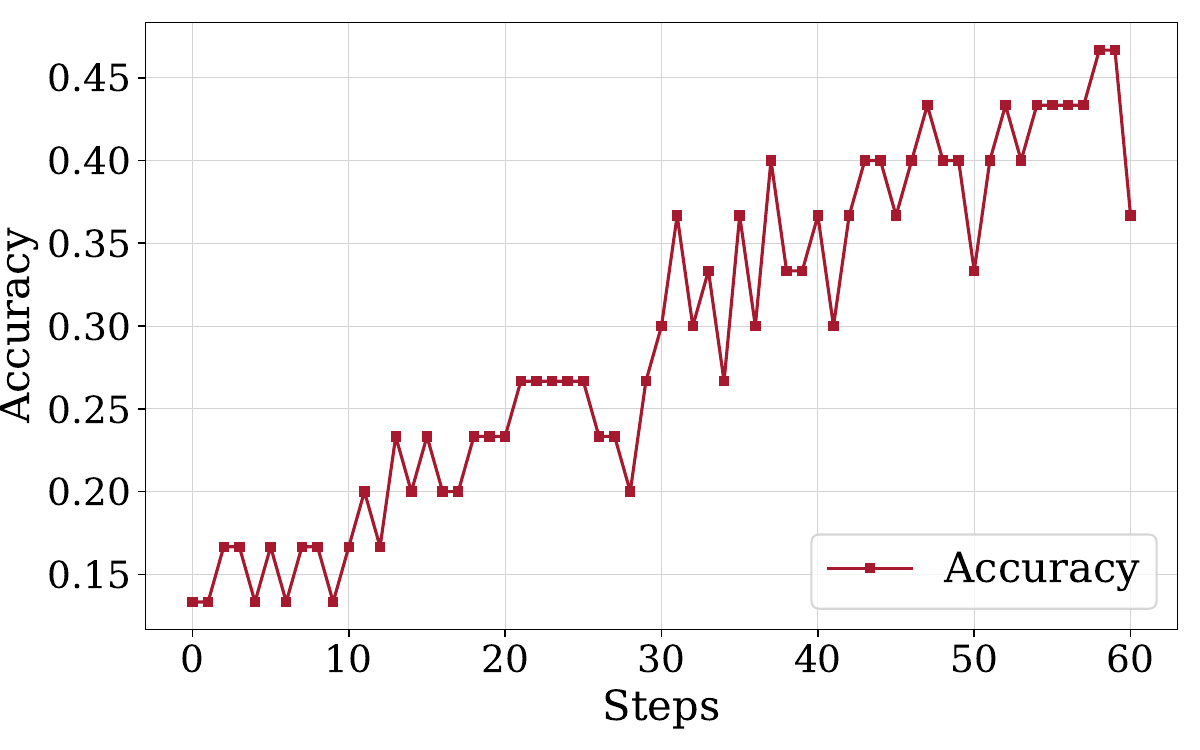}
        \caption{Accuracy Curve.}
        \label{fig:aime_acc}
    \end{subfigure}
    \hfill
    \begin{subfigure}[t]{0.48\textwidth}
        \centering
        \includegraphics[width=\linewidth]{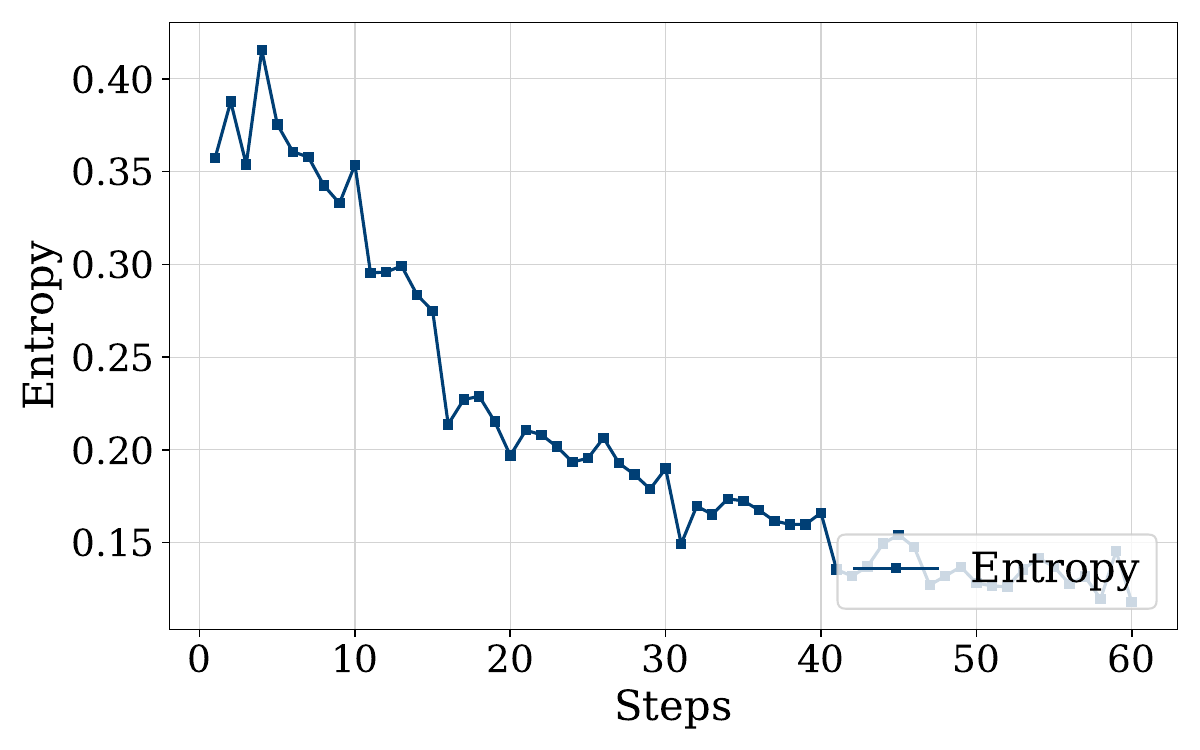}
        \caption{Entropy Curve.}
        \label{fig:aime_entropy}
    \end{subfigure}
    \caption{The entropy and accuracy curves of \method on AIME 2024 with \texttt{Qwen2.5-Math-7B}.
    }
    \label{fig:aime_curves}
\end{figure}

On the highly challenging mathematical reasoning benchmark AIME 2024, \method achieves a substantial improvement of $159.3\%$, surpassing all models trained on large-scale datasets.
Furthermore, when applied to \texttt{Qwen2.5-Math-7B}, \method yields an average improvement of $84.1\%$ across three benchmarks.
Figure~\ref{fig:aime_curves} shows two curves of \method{} on AIME 2024 with \texttt{Qwen2.5-Math-7B} as an example.

\begin{table}[!h]
\centering
\caption{Additional results of \method{} on each task. $^*$ indicates results from Dr. GRPO~\citep{liu2025understanding}.
Our training data size matches the corresponding benchmark dataset size.
}
\label{tab:additional-results}
\resizebox{\linewidth}{!}{
\begin{tabular}{lccccc}
\toprule
\textbf{Name} & \textbf{AIME 2024} & \textbf{AMC} & \textbf{MATH-500} & \textbf{Avg} & \textbf{Labeled Data} \\ \midrule
Qwen2.5-Math-1.5B$^*$ & $20.0$ & $32.5$ & $33.0$ & $28.5$ & - \\ \midrule
\rowcolor{lightblue!100}\multicolumn{1}{r}{\textbf{w/ \method}} & $20.0$ & $53.0$ & $\mathbf{80.0}$ & $\mathbf{51.0}$ & \xmark \\
\rowcolor{lightblue!100}\multicolumn{1}{r}{$\Delta$} & $0$ & $+20.5$ & $+47.0$ & $+22.5$ & \xmark \\
& $0$ & \textcolor{red}{$\uparrow63.1\%$} & \textcolor{red}{$\uparrow142.4\%$} & \textcolor{red}{$\uparrow79.0\%$} & \xmark \\
\midrule
\midrule
\textcolor{gray}{Qwen2.5-Math-1.5B-Instruct$^*$} & \textcolor{gray}{$10.0$} & \textcolor{gray}{$48.2$} & \textcolor{gray}{$74.2$} & \textcolor{gray}{$44.1$} & \textcolor{gray}{$3.1$M} \\
\textcolor{gray}{DeepSeek-R1-Distill-1.5B@3k$^*$} & \textcolor{gray}{$2.5$} & \textcolor{gray}{$21.7$} & \textcolor{gray}{$52.2$} & \textcolor{gray}{$25.5$} & \textcolor{gray}{$800$K} \\
\textcolor{gray}{DeepSeek-R1-Distill-1.5B@8k$^*$} & \textcolor{gray}{$20.0$} & \textcolor{gray}{$49.4$} & \textcolor{gray}{$77.4$} & \textcolor{gray}{$48.9$} & \textcolor{gray}{$800$K} \\
\textcolor{gray}{Oat-Zero-1.5B$^*$} & \textcolor{gray}{$\mathbf{20.0}$} & \textcolor{gray}{$\mathbf{53.0}$} & \textcolor{gray}{$74.2$} & \textcolor{gray}{$49.1$} & \textcolor{gray}{$8.9$K} \\
\toprule
Qwen2.5-Math-7B$^*$ & $16.7$ & $38.6$ & $50.6$ & $35.3$ & - \\
\midrule
\rowcolor{lightblue!100}\multicolumn{1}{r}{\textbf{w/ \method}} & $43.3$ & $\mathbf{67.5}$ & $\mathbf{84.2}$ & $\mathbf{65.0}$ & \xmark \\
\rowcolor{lightblue!100}\multicolumn{1}{r}{$\Delta$} & $+26.6$ & $+28.9$ & $+33.6$ & $+29.7$ & \xmark \\
& \textcolor{red}{$\uparrow159.3\%$} & \textcolor{red}{$\uparrow74.9\%$} & \textcolor{red}{$\uparrow66.4\%$} & \textcolor{red}{$\uparrow84.1\%$} & \xmark \\
\midrule
\midrule
\textcolor{gray}{Qwen2.5-Math-7B-Instruct$^*$} & \textcolor{gray}{$16.7$} & \textcolor{gray}{$53.0$} & \textcolor{gray}{$83.6$} & \textcolor{gray}{$51.1$} & \textcolor{gray}{$3.1$M} \\
\textcolor{gray}{DeepSeek-R1-Distill-7B@3k$^*$} & \textcolor{gray}{$10.0$} & \textcolor{gray}{$26.2$} & \textcolor{gray}{$60.1$} & \textcolor{gray}{$32.1$} & \textcolor{gray}{$800$K} \\
\textcolor{gray}{SimpleRL-Zero-7B$^*$} & \textcolor{gray}{$26.7$} & \textcolor{gray}{$60.2$} & \textcolor{gray}{$78.2$} & \textcolor{gray}{$55.0$} & \textcolor{gray}{$8.9$K} \\
\textcolor{gray}{PRIME-Zero-7B$^*$} & \textcolor{gray}{$16.7$} & \textcolor{gray}{$62.7$} & \textcolor{gray}{$83.8$} & \textcolor{gray}{$54.4$} & \textcolor{gray}{$230$K} \\
\textcolor{gray}{OpenReasoner-Zero-7B@3k$^*$} & \textcolor{gray}{$13.3$} & \textcolor{gray}{$47.0$} & \textcolor{gray}{$79.2$} & \textcolor{gray}{$46.5$} & \textcolor{gray}{$129$K} \\
\textcolor{gray}{Oat-Zero-7B$^*$} & \textcolor{gray}{$\mathbf{43.3}$} & \textcolor{gray}{$62.7$} & \textcolor{gray}{$80.0$} & \textcolor{gray}{$62.0$} & \textcolor{gray}{$8.9$K} \\
\textcolor{gray}{LIMR-7B} & \textcolor{gray}{$32.5$} & \textcolor{gray}{$63.8$} & \textcolor{gray}{$78.0$} & \textcolor{gray}{$58.1$} & \textcolor{gray}{$1.4$K} \\
\bottomrule
\end{tabular}
}
\end{table}

\section{Training Metrics}\label{appx:training_metrics}
Given the absence of ground-truth labels in the test data, evaluating the performance of \method throughout the training process presents a challenge.
To mitigate this limitation, we introduce a set of training-time metrics specifically designed to monitor and assess the effectiveness of \method.
These metrics inform the selection of the optimal checkpoint and provide valuable insights regarding training dynamics.

\begin{itemize}[leftmargin=1em]

\item \textbf{Entropy}: Measures the uncertainty of the model’s generation.

\item \textbf{Majority Voting Reward}: Rule-based rewards computed from the majority-voted label.

\item \textbf{Majority Ratio}: The frequency of the most common answer within a rollout.

\end{itemize}

Furthermore, we define several metrics that rely on access to ground-truth labels, which allow for a deeper analysis of the model’s behavior during training:

\begin{itemize}[leftmargin=1em]

\item \textbf{Label Accuracy (maj@n)}: Indicates whether the estimated label matches ground-truth.

\item \textbf{Reward Accuracy}: Indicates the proportion of majority voting rewards (computed from the estimated label) that match rewards computed from the ground-truth label.

\item \textbf{Ground-Truth Ratio}: The frequency of the ground-truth answer within a rollout.

\end{itemize}

\section{Terminology}
\label{sec:appendix_terminology}
Test-time scaling refers to increasing computational resources during test time, which can be categorized into test-time training and test-time inference. These two approaches are complementary. We will provide an introduction below.

\begin{table}[!ht]
\centering
\caption{Terminology relationship.}
\label{tab:tts-ttrl}
\resizebox{\linewidth}{!}{
\begin{tabular}{@{}ccc@{}}
\toprule
Name & Category & Methods \\ \midrule
\multirow{2}{*}{Test-Time Scaling (TTS)} 
     & Test-Time Training (TTT)     & Test-Time Reinforcement Learning (TTRL)  \\ 
     & Test-Time Inference (TTI) & Majority Voting, Best-of-N \\\bottomrule
\end{tabular}
}
\end{table}

\subsection{Test-Time Training (TTT)}
Test-Time Training (TTT) is a technique for adapting a pre-trained model at inference time to improve generalization under distribution shifts. Let $f_\theta$ denote a model trained on a source domain $\mathcal{D}s = \{(x_i, y_i)\}{i=1}^N$, where $x_i \in \mathcal{X}$, $y_i \in \mathcal{Y}$, and $\theta$ represents the learned parameters. During standard inference, the model is evaluated on test samples $x_t \sim \mathcal{D}_t$ with fixed parameters $\theta$, where $\mathcal{D}_t \neq \mathcal{D}_s$.

In contrast, TTT allows the model to adapt to each test sample $x_t$ by minimizing an auxiliary self-supervised loss $\mathcal{L}_{\text{aux}}$, without access to labels $y_t$. The model parameters are updated online with the auxiliary task, which is typically designed to be label-free and consistent with the main task.

\subsection{Test-Time Inference (TTI)}

Test-Time Inference (TTI) refers to the strategy of enhancing the performance of a large language model during inference by allocating additional computational resources.
Formally, let $f_\theta$ denote a language model with parameters $\theta$, and let $x$ be an input prompt.
The model generates an output $y$ by sampling from the conditional distribution $p_\theta(y \mid x)$. TTI techniques aim to improve the quality of $y$ by employing methods such as generating multiple candidate outputs and selecting the best one based on a scoring function, or by refining the output through iterative processes~\citep{welleck2024decoding}.

One common approach involves generating $N$ candidate outputs $\{y_1, y_2, \ldots, y_N\}$ and selecting the optimal output $y^*$ using a scoring function $s(y, x)$:

\begin{align}
y^* = \arg\max_{y_i} s(y_i, x)
  \label{eq:tti}
\end{align}

The scoring function $s(y, x)$ can be instantiated in various ways, such as:
\begin{enumerate}[leftmargin=2em]
    \item Majority Voting (MV): Selecting the most frequent output among the candidates.
    \item Best-of-N (BoN): Using reward models to score each candidate, then selecting the highest-scoring one.
    \item Weighted BoN: Integrating MV and BoN strategies to leverage their respective strengths.
\end{enumerate}

\end{document}